\documentclass[letterpaper, 10 pt, conference]{ieeeconf}  

\IEEEoverridecommandlockouts                              

\usepackage{graphicx}
\usepackage{amsmath,amssymb}
\usepackage{multirow}
\usepackage{float}
\usepackage{makecell}
\usepackage{booktabs}
\usepackage{tabularx}
\usepackage{caption}
\usepackage{url}
\usepackage{pifont}
\usepackage[hidelinks]{hyperref}
\hypersetup{
  pdftitle={AetheRock: An Arm-Worn Robot Teaching System for Force-Guided Vision-Tactile Learning},
  pdfauthor={Hong Li, Yue Xu, Yihan Tang, Yankang Dong, Chenyuan Liu, Chenyang Yu, Xuyang Li, Siyuan Huang, Yujun Shen, Nan Xue, Yong-Lu Li},
  pdfkeywords={Force-Tactile Sensor, Robot System, Contact-rich Manipulation}
}

\title{\LARGE \bf
AetheRock: An Arm-Worn Robot Teaching System for Force-Guided Vision-Tactile Learning
}
\author{
\authorblockN{Hong Li\textsuperscript{1}, Yue Xu\textsuperscript{1},
Yihan Tang\textsuperscript{1}, Yankang Dong\textsuperscript{1},
Chenyuan Liu\textsuperscript{1}, Chenyang Yu\textsuperscript{1},\\
Xuyang Li\textsuperscript{1,*}\thanks{\textsuperscript{*}This work was done during an internship at Shanghai Jiao Tong University.},
Siyuan Huang\textsuperscript{4}, Yujun Shen\textsuperscript{2},
Nan Xue\textsuperscript{2}, Yong-Lu Li\textsuperscript{1,3,\ensuremath{\dagger}}\thanks{\textsuperscript{\ensuremath{\dagger}}Correspondence to: Yong-Lu Li \texttt{<yonglu\_li@sjtu.edu.cn>}.}}
\authorblockA{\textsuperscript{1}Shanghai Jiao Tong University,
\textsuperscript{2}Ant Group,
\textsuperscript{3}Shanghai Innovation Institute\\
\textsuperscript{4}Beijing Institute for General Artificial Intelligence (BIGAI)\\
\texttt{\{hong\_li, yonglu\_li\}@sjtu.edu.cn}}
}

\begin{document}

\IEEEaftertitletext{%
    \begin{minipage}{\textwidth}
        \centering
        \includegraphics[width=0.95\linewidth]{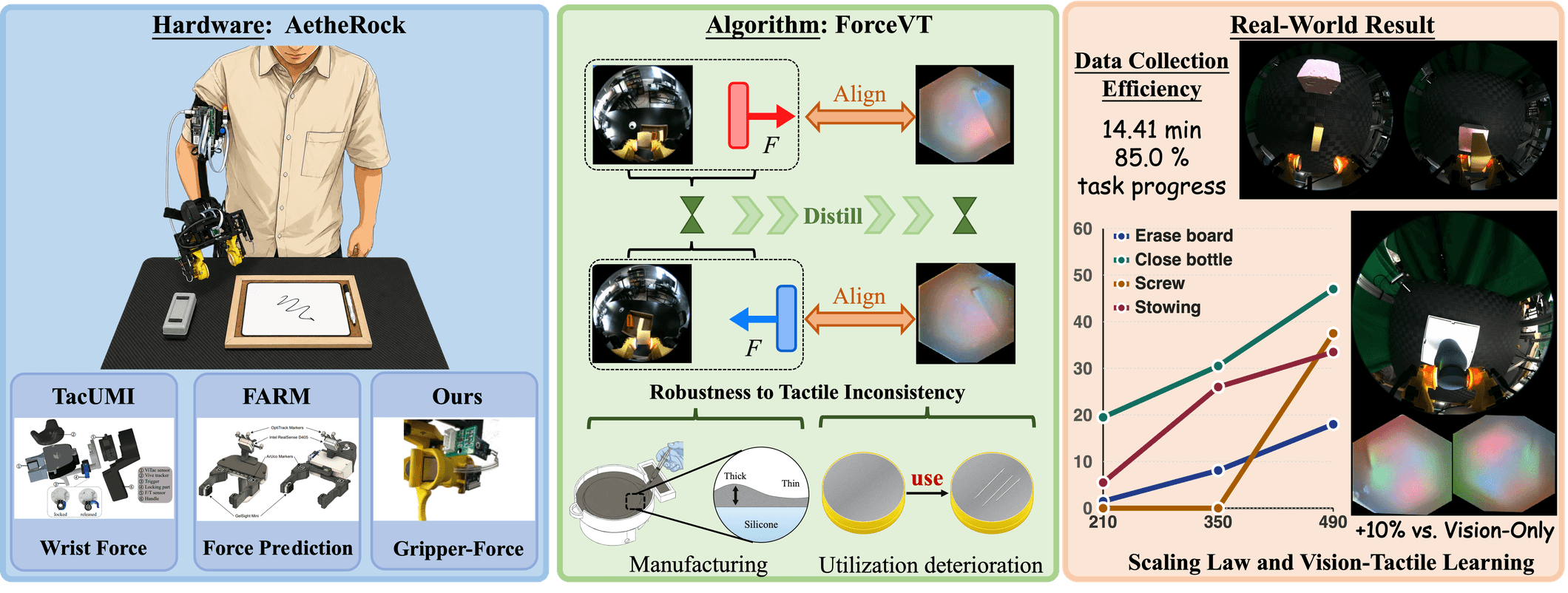}
        \captionof{figure}{We present a hardware and algorithm co-design robot teaching system. We introduce AetheRock (Left), an arm-worn device for collecting gripper force, vision, and tactile data. We further develop the ForceVT algorithm (Middle) to improve robust tactile utilization. We evaluate the effectiveness of the hardware and algorithm in real-world robotics (Right).}
        \label{fig:teaser}
    \end{minipage}
    \vspace{0.5\baselineskip}
}
\maketitle
\thispagestyle{empty}
\pagestyle{empty}

\begin{abstract}
Force and tactile sensing are essential for contact-rich manipulation, but their integration into handheld or wearable devices is constrained by cost, size, and mechanical compatibility. Tactile signals are also sensitive to manufacturing variations and wear during use, posing further challenges to policy learning. To address these hardware and learning challenges, we introduce \textit{AetheRock}, an arm-worn device for collecting gripper-force, visual, and tactile data. AetheRock incorporates \textit{GelSlim-Mini}, our improved version of GelSlim 4.0~\cite{sipos2024gelslim} designed for low-cost, easy manufacturing, at the gripper fingertip, together with a resistive pressure sensor at the human finger contact area. To learn robust representations from the collected data, we propose \textit{ForceVT}, a framework that aligns multimodal representations and uses force and vision to guide tactile learning, mitigating the effects of manufacturing variations and sensor degradation. Real-world experiments demonstrate efficient data collection with AetheRock. ForceVT improves average performance by 13\% over the vision-tactile baseline and maintains an average task progress of 55.0\% under reduced tactile fidelity, compared with 58.0\% at full fidelity.
\par
\begingroup
\setlength{\parskip}{3pt}
\nobreak\vskip 3pt
\noindent\mbox{\normalfont\fontsize{11}{13}\selectfont
  \textbf{Project page:}\enspace\urlstyle{same}\url{https://lihongcs.github.io/aetherock}.}\par
\endgroup
\end{abstract}

\section{Introduction}

\begin{figure*}[!t] \centering
\includegraphics[width=0.9\linewidth]{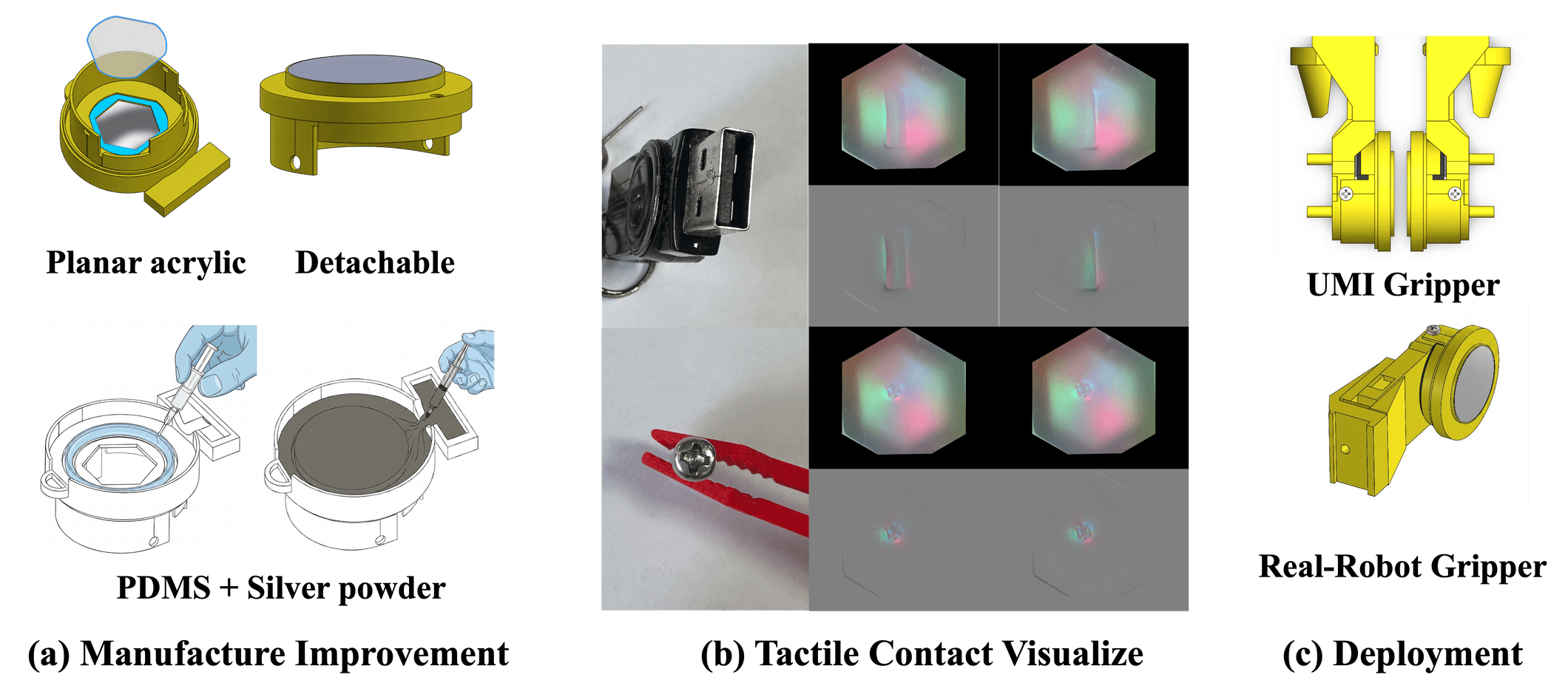}
\caption{Manufacturing of GelSlim-Mini. Inspired by the GelSlim 4.0~\cite{sipos2024gelslim} optics and PCB design, we propose GelSlim-Mini, including low-cost manufacturing, easy repairability, and adaptability to multiple gripper features.}
\label{fig:gelslim-Mini}
\end{figure*} Recent advances in robot learning have driven substantial progress in imitation learning~\cite{chi2025diffusion} and vision-language-action (VLA)~\cite{black2024pi_0,intelligence2025pi_} models. These are driven by large-scale robot manipulation data. While teleoperation yields the highest benefit in robot learning, it is labor-intensive and scene-limited. To reduce the complexity of data acquisition, many works involve egocentric human demonstrations~\cite{punamiya2026egoverse} and UMI~\cite{chi2024universal, choi2026wild} data. Human demonstrations are cheaper and more massive, but exhibit low adaptability to robot embodiments. UMI, with the same end effector and relative trajectory movement as a real robot, shows huge potential for collecting robot teaching data.

Force and tactile sensing are essential for contact-rich manipulation, motivating more studies to incorporate them into data collection systems. To acquire force signals, some systems~\cite{cheng2026tacumi,luo2026omniumi} mount 6D F/T force sensors between the gripper and a rear handle to measure wrist forces. However, their high cost limits scalability, and their large size restricts installation at desired locations. Other methods, such as FARM~\cite{helmut2025tactile}, estimate force from visuo-tactile signals, but their accuracy is constrained by the estimation algorithm and tactile sensing reliability. In contrast, visuo-tactile sensors provide rich contact information and are easily integrated into grippers~\cite{xu2025exumi, zhu2026touch}, but suffer from manufacturing inconsistencies and degradation during use. Variations in gel mixing and coating introduce manufacturing inconsistencies, while gel-edge detachment and impact damage during use gradually alter sensing fidelity and sensitivity. These issues can degrade policy performance.

To address these challenges, we present \textbf{AetheRock}, an arm-worn robot teaching system for collecting gripper-force, visual, and tactile data, as shown on the left of Figure~\ref{fig:teaser}. Through finger-operated control, our system collects UMI end-effector data together with direct gripper-force measurements and tactile observations, offering a low-cost alternative to designs using 6D force/torque sensors or force estimation. Specifically, our system introduces \textbf{two key innovations}: a \textbf{resistive pressure sensor} integrated with the central controller for robust gripper-force acquisition, and \textbf{GelSlim-Mini}, a modified version of the GelSlim 4.0~\cite{sipos2024gelslim} visuo-tactile sensor designed for low-cost, easy manufacturing.

Building on AetheRock, we propose \textbf{ForceVT}, a force-guided vision-tactile learning algorithm, as shown in middle of Figure~\ref{fig:teaser}. It combines multimodal alignment with force-guided tactile learning to improve cross-modal representations and mitigate the effects of visuo-tactile inconsistencies. To study these effects, we construct tactile sensors with varying sensing fidelities to simulate manufacturing variations and degradation during use, collecting synchronized gripper-force, tactile, and fisheye visual signals under each configuration. Our experiments, summarized on the right of Figure~\ref{fig:teaser}, evaluate data collection efficiency, the benefits of force and tactile sensing, and policy robustness to tactile sensing inconsistencies. Specifically, we assess collection efficiency on six robotic tasks and investigate post-training scaling laws and vision-tactile learning on four tasks through comparisons with vision-only baselines and evaluations under varying tactile fidelities. The results demonstrate that AetheRock provides high-quality demonstration data, force-tactile information improves task performance, and ForceVT enhances robustness to tactile sensing inconsistencies.

In summary, our primary contributions include: \textbf{(1)} An arm-worn robot teaching system, AetheRock, for collecting gripper-force, visual, and tactile robot data. \textbf{(2)} An improved visuo-tactile sensor, GelSlim-Mini, designed for low cost and easy manufacturing, and a mechanical assembly incorporating a resistive pressure sensor for gripper-force collection. \textbf{(3)} A force-guided vision-tactile learning algorithm, ForceVT, that uses force and vision to guide fidelity-agnostic tactile representation learning, improving performance by 13\% over the vision-tactile baseline and maintaining an average task progress of 55.0\% under reduced tactile fidelity, compared with 58.0\% at full fidelity.

\section{Related Work}

\begin{table}[!htb] 
\centering 
\caption{Comparison of Sensor fabrication.} \label{tab:tactile_hardware_comparison} \fontsize{8.5}{10}\selectfont \setlength{\tabcolsep}{3pt} \renewcommand{\arraystretch}{1.25} \resizebox{\linewidth}{!}{\begin{tabular}{@{}lccccc@{}} \toprule Sensor & Planar acrylic & Elastomer & Coating & Replaceable & Cost (USD) \\ \midrule GelSlim 4.0~\cite{sipos2024gelslim} & \ding{55} & XP-565 & Silicone ink & \ding{55} & $\sim$65 \\ Ours & \ding{51} & PDMS & Silver powder & \ding{51} & $\sim$10 \\ \bottomrule \end{tabular}} 
\end{table} 
\begin{figure*}[!t]
    \centering
    \includegraphics[width=0.95\linewidth]{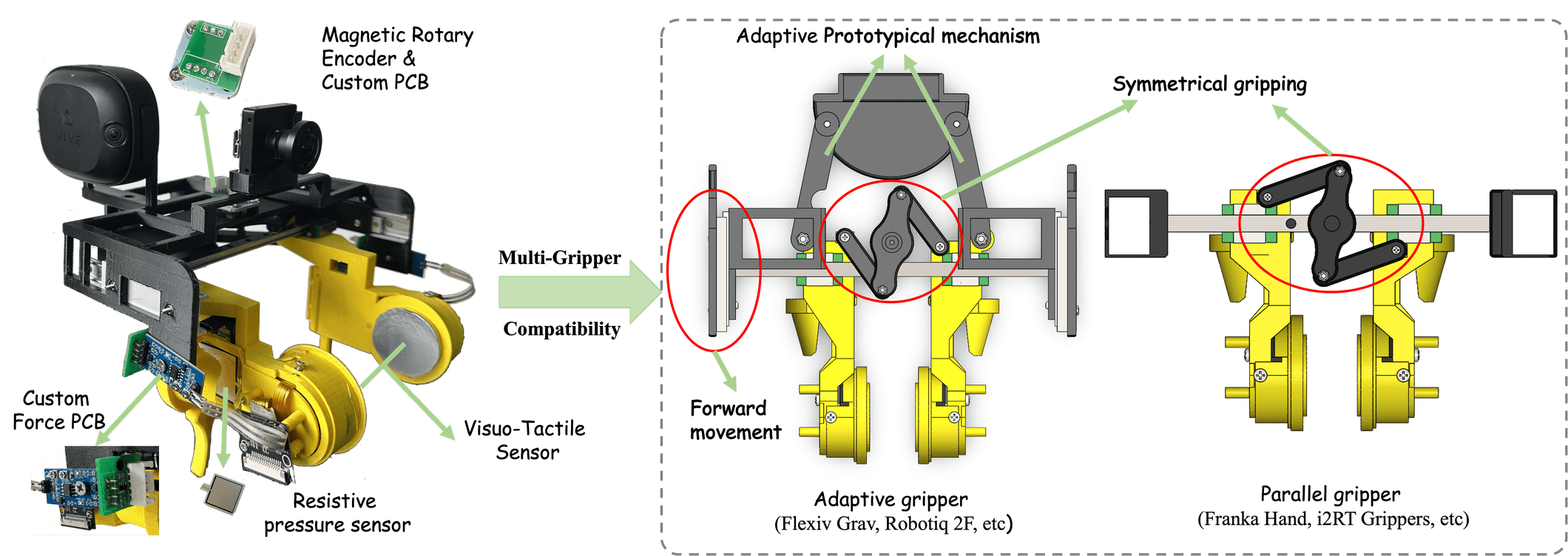}
    \caption{AetheRock hardware system. An arm-worn data collection system for force, vision, and tactile data collection, which includes two general grippers, an adaptive gripper and a parallel gripper, can be easily adapted to other robot grippers.}
    \label{fig:hardware}
\end{figure*} \paragraph{Robot Data Collection System.} Universal Manipulation Interface(UMI) shares the same end-effector as the real robot and learns using the relative movement of the executor, which exhibits great potential for general robot manipulation models. FastUMI~\cite{zhaxizhuoma2025fastumi} improves the SLAM system with RealSense T265. RoboPocket~\cite{fang2026robopocket} integrates its system within an iPhone. exUMI~\cite{xu2025exumi}, FreeTacMan~\cite{wu2025freetacman},  TouchGuide~\cite{zhang2026touchguide}, TacUMI~\cite{cheng2026tacumi} incorporate visuo-tactile sensors. ForceMimic~\cite{liu2025forcemimic} and OmniUMI~\cite{luo2026omniumi} integrate 6-axis F/T sensors.
Although existing works integrate more sensors, gripper force is not considered in current works, which is significant for contact-rich tasks. We propose an Arm-Worn Robot Teaching System, AetheRock, to collect force, vision and tactile data.

\paragraph{Visuo-Tactile Sensor Design.} Various tactile sensors are designed for contact-rich manipulation, including visuo-tactile~\cite{lin20239dtact,lambeta2020digit,yuan2017gelsight}, capacitative~\cite{glauser2019deformation, wu2020capacitivo, xu2024cushsense}, resistive~\cite{sundaram2019learning, stassi2014flexible}, etc. Visuo-tactile sensors are widely applied, offering ease of fabrication and low cost. However, all tactile sensors have a limited lifecycle, and their gel are easily destroyed. GelSlim 4.0~\cite{sipos2024gelslim} is an open-source tactile sensor. Although it provides complete manufacturing details, the dependent devices are high-cost, and the repair process is tedious. To address these challenges and provide a hardware base for algorithms resolving manufacturing and utilization inconsistencies, we propose GelSlim-Mini.

\paragraph{Vision-Tactile Learning.} Vision-tactile learning focuses on training a tactile encoder by aligning tactile features with vision features or injecting tactile tokens into vision features. More closely related to robot action models, many works aim to incorporate tactile tokens into VLA models. VTLA~\cite{zhang2026vtla} trains a Vision-Tactile-Language-Action model by integrating vision and tactile inputs through cross-modal language grounding. OmniVTLA~\cite{cheng2025omnivtla} utilizes a dual-path tactile framework for the encoder and semantically aligned algorithms between language and vision. TacVLA~\cite{zhang2026tacvla} and AT-VLA~\cite{li2026vla} employ a gating mechanism to achieve adaptive multimodal fusion. Tactile-VLA~\cite{huang2025tactile} divides the model’s intentions into precise physical actions and a reasoning module to obtain vision-tactile perceptions. HapticVLA~\cite{gubernatorov2026hapticvla} studies training with offline data involving haptics and deployment without direct haptic feedback during inference. TacFiLM~\cite{morissette2026tactile} extends FiLM~\cite{perez2018film} fusion to vision-tactile learning. Overall, these works rely on ideal tactile sensors, overlooking visuo-tactile inconsistencies caused by manufacturing and utilization. Therefore, we propose an algorithm to study the robust utilization of tactile sensors.

\section{AetheRock Data Collection System}

\begin{figure*}[!t]
    \centering
    \includegraphics[width=1.0\linewidth]{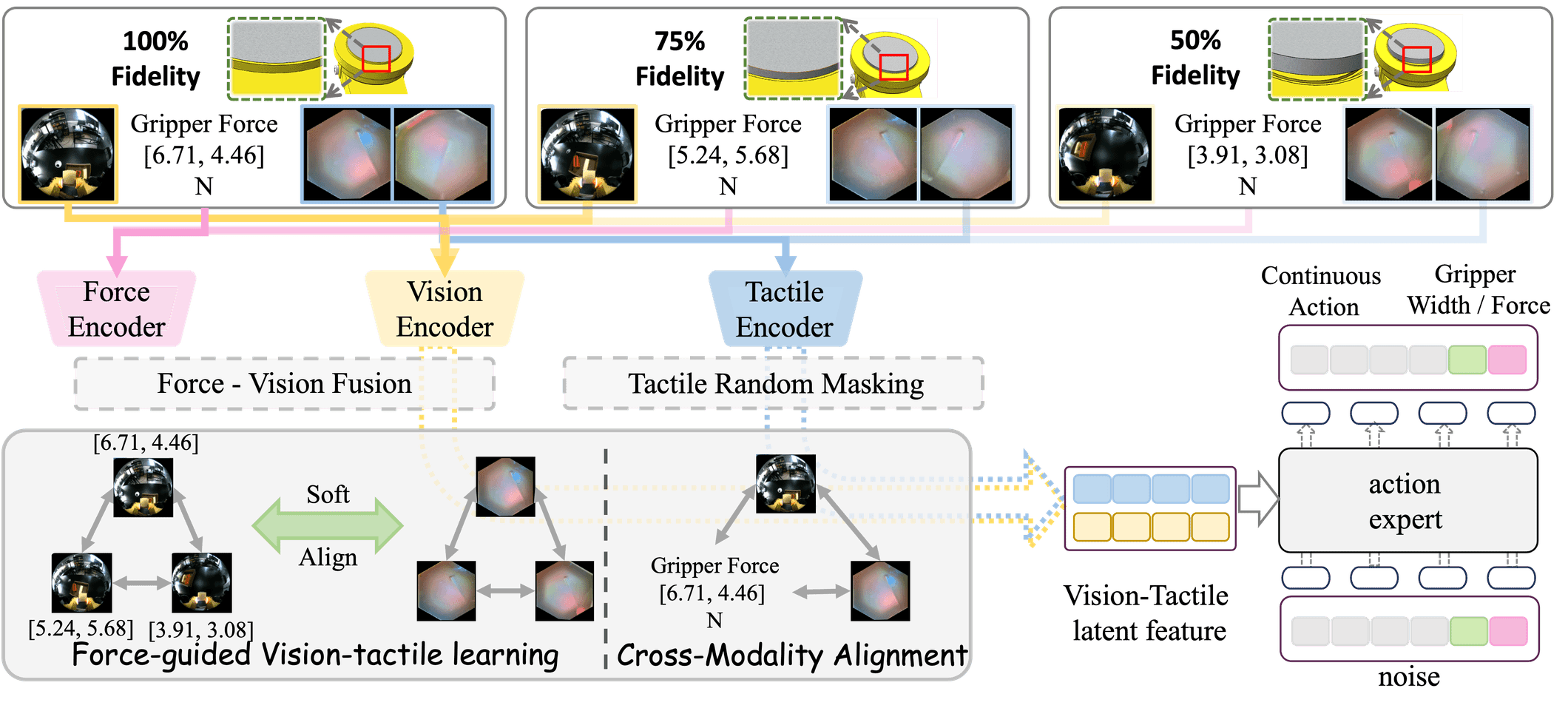}
    \caption{Schematic framework of the proposed ForceVT. Multi-fidelity tactile data are augmented via random masking, followed by modality-specific encoding. The force-guided vision-tactile learning module leverages fused vision-force features to guide tactile representation learning. Finally, the cross-modality alignment module computes cosine similarities across all modalities, while the conditioned vision-tactile features generate continuous actions.}
    \label{fig:methods}
\end{figure*} 

\subsection{Hardware Design}

\paragraph{Gelslim-Mini} Inspired by the optics and PCB design of GelSlim 4.0~\cite{sipos2024gelslim}, we propose \textbf{GelSlim-Mini}, a low-cost, easy-to-manufacture visuo-tactile sensor, as shown in Figure~\ref{fig:gelslim-Mini} and Table~\ref{tab:tactile_hardware_comparison}. GelSlim-Mini offers three key improvements over GelSlim 4.0: (1) \textbf{Planar acrylic}, allowing the use of 2D-cut acrylic instead of costly 3D acrylic. (2) \textbf{Detachable components}, enabling quick repair of damaged RGB components and silicone cases. (3) \textbf{Low cost}, providing clear tactile signals at approximately one-sixth of the cost, as shown in Figure~\ref{fig:gelslim-Mini}(b). In our experiments, the sensors are mounted on robot grippers and UMI devices, as shown in Figure~\ref{fig:gelslim-Mini}(c).

\paragraph{Multi-sensor Data Collection Module} As shown in Figure~\ref{fig:hardware}, for visual sensing, we follow RDT2~\cite{liu2026rdt2} and use a Hikrobot MV-CB060-10UM/UC industrial camera with an exposure of 20,000. For motion tracking, we adopt an HTC VIVE Ultimate Tracker and adapt it to a headless setup following DexCap~\cite{wang2024dexcap}. Following exUMI~\cite{xu2025exumi}, gripper width is estimated using a low-cost AS5600 magnetic rotary encoder with an improved PCB design for compactness and robustness. To improve the generalizability of AetheRock, we develop both parallel and adaptive gripper designs, which can generalize to other grippers. To avoid the size and cost limitations of 6D F/T sensors, we use a low-cost resistive pressure sensor mounted at human finger contact regions for gripper-force collection.

\subsection{Data Process}

\paragraph{System Time Calibration.} To guarantee precise temporal alignment across the PC and OrangePi central control during data collection, we developed a localized high-precision Network Time Protocol (NTP) synchronization mechanism. The PC acts as the primary time server, while the OrangePi controller functions as a dedicated client. By enforcing high-frequency local polling for frequent synchronization and isolating it from external public time pools, we achieve highly stable temporal alignment across all sensor streams with an average synchronization latency of $10\text{ ms}$.

\paragraph{Tactile Calibration.} To ensure spatial consistency and eliminate perspective distortion in the tactile observations, we develop a geometric calibration pipeline. First, the gel region is isolated from the surrounding support frame using adaptive thresholding and polygonal approximation. Next, we reconstruct the complete sensor geometry by extrapolating the valid contour edges to recover the hidden vertices, effectively capturing the inherent perspective tilt and spatial asymmetry. Finally, a homography transformation maps these distorted physical vertices onto a dynamically calculated, mathematically ideal regular polygon centered within the image frame. This spatial warping process yields a standardized tactile vision, eliminating geometric inconsistencies across hardware devices.

\paragraph{Resistive Pressure Sensor Calibration.} To process the resistive pressure sensor readings, we implement a piecewise linear calibration protocol via a 16-bit I2C ADC interface. The raw analog signals are first mapped to the system voltage domain. To mitigate sensor noise and hardware saturation, we apply a thresholding mechanism: voltages below an effective threshold of $500\text{ mV}$ are zeroed, while the active response range ($500$--$5000\text{ mV}$) is linearly mapped to the target physical force span ($50$--$1000\text{ g}$). The final measurements are converted to standard Newtons ($\text{N}$).

\section{Methodology}

\subsection{Robust Tactile Representation Learning}

To increase tactile diversity and improve robustness to any tactile situation, we introduce \textbf{Tactile Random Masking}, which perturbs tactile observations through stochastic spatial masking. Following DINOv3~\cite{simeoni2025dinov3}, for each training batch, a ratio $\rho \in (0,1)$ of tactile samples is randomly selected for masking, where a fraction $\eta \in (0,1)$ of spatial patches in each selected sample is randomly replaced. For tactile observations from the $k$-th fidelity sensor, the process is formulated as:
\begin{equation}
    \tilde{T}^{(k)} = \mathcal{M}_{\rho,\eta}(T^{(k)}),
\end{equation}
where $\mathcal{M}_{\rho,\eta}$ denotes the masking operator, and $T^{(k)}$ and $\tilde{T}^{(k)}$ represent the raw and masked tactile observations from the $k$-th fidelity sensor, respectively.

The masked tactile input $\tilde{T}^{(k)}$, together with aligned wrist vision $I_v$ and force data $F$, is encoded by modality-specific encoders to obtain latent representations:
\begin{equation}
    z_v = E_v(I_v), \quad z_f = E_f(F), \quad \tilde{z}_t^{(k)} = E_t(\tilde{T}^{(k)}),
\end{equation}
where $E_v$, $E_f$, and $E_t$ denote the vision, force, and tactile encoders, respectively.

\subsection{ForceVT: Force Guided Vision Tactile Learning}

\begin{figure*}[!t]
\centering
\begin{minipage}[t]{0.31\linewidth}
\vspace{0pt}
\centering
\includegraphics[width=\linewidth]{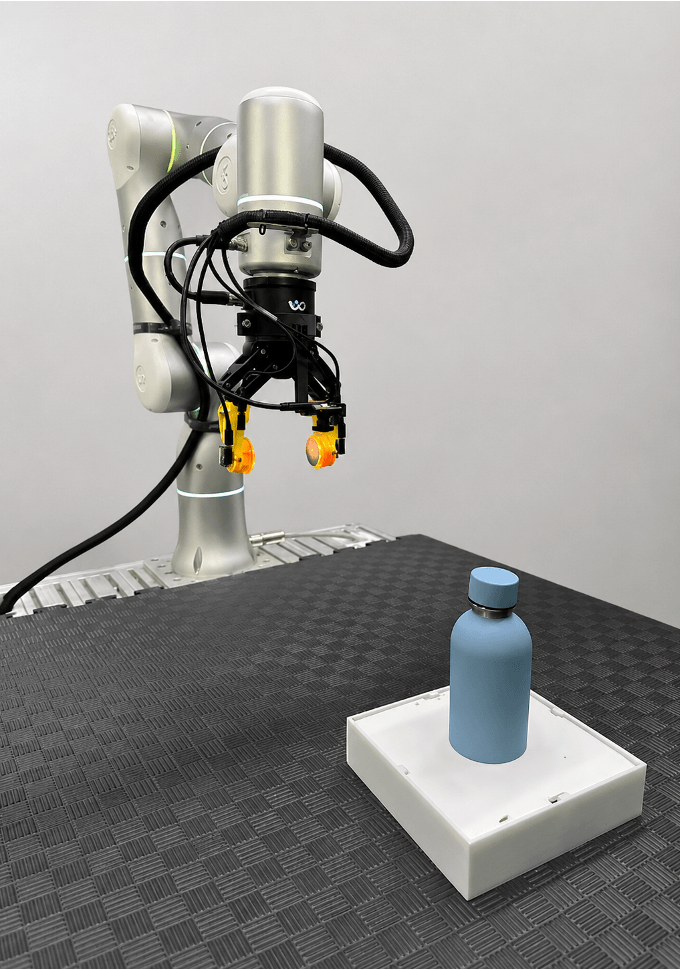}
\caption{Real-robot experimental setup.}
\label{fig:real_robot_setup}
\end{minipage}
\hfill
\begin{minipage}[t]{0.66\linewidth}
\vspace{0pt}
    \centering
    \includegraphics[width=\linewidth]{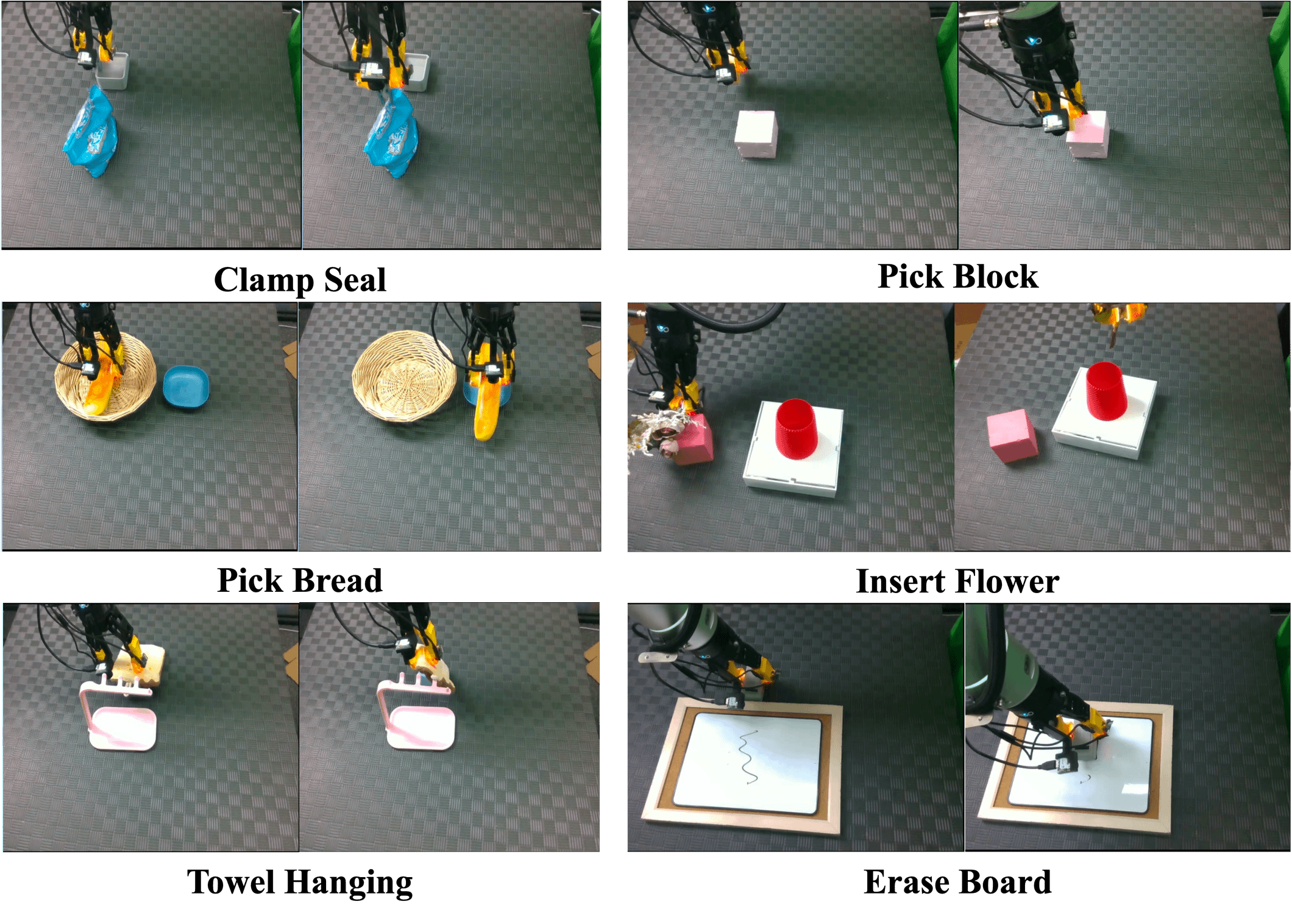}
    \caption{Real-world rollouts across six tasks using policies with fisheye wrist camera images as the input.}
    \label{fig:single_generation}
\end{minipage}\end{figure*} 
Based on our proposed visuo-tactile sensor, GelSlim-Mini, we fabricate sensors across various fidelities to collect aligned wrist vision, tactile and gripper force data. Furthermore, we propose a force-guided vision-tactile learning algorithm to learn fidelity-agnostic tactile representations, ensuring robustness in any tactile situation. 
Figure~\ref{fig:methods} presents ForceVT, a post-training framework featuring \textbf{Robust Tactile Representation Learning} for robustness to tactile variations and \textbf{Cross-Modality Soft Alignment} for preserving consistency between the relational structure of tactile features and that of fused vision–force features.

We first fuse vision and force features with a lightweight fusion module:
\begin{equation}
z_{vf} = \mathcal{F}(z_v, z_f),
\end{equation}
where $z_{vf}$ denotes the fused vision-force representation.

To learn fidelity-agnostic tactile representations, we introduce a soft alignment that transfers relational similarity across different tactile fidelities from the fused vision force space to tactile space:
\begin{equation}
\mathcal{L}_{\text{soft}}
=
\frac{1}{N^2}
\sum_{i,j}
\left\|
\mathrm{sim}(z_{vf}^i, z_{vf}^j)
-
\mathrm{sim}(\tilde{z}_{t}^{i,(k_i)}, \tilde{z}_{t}^{j,(k_j)})
\right\|_2^2,
\end{equation}
where $k_i$ and $k_j$ denote the tactile fidelity levels of samples $i$ and $j$, respectively. Thus, the tactile similarity is computed between tactile representations that may come from different fidelity sensors.

To achieve cross-modality alignment, we adopt a contrastive loss between vision force and tactile representations:
\begin{equation}
\mathcal{L}_{\text{con}}
=
-\frac{1}{N}
\sum_{i=1}^{N}
\log
\frac{
\exp(\mathrm{sim}(z^i_{vf}, \tilde{z}^{i,(k_i)}_t)/\tau)
}{
\sum_{j=1}^{N}
\exp(\mathrm{sim}(z^i_{vf}, \tilde{z}^{j,(k_j)}_t)/\tau)
},
\end{equation}
where $\mathrm{sim}(\cdot,\cdot)$ denotes cosine similarity, $\tau$ is a temperature parameter, and $N$ is the batch size.

During inference, the policy predicts continuous actions $a$, including end-effector relative pose, gripper width, and force, from vision and tactile features. The action objective is defined as:
\begin{equation}
\mathcal{L}_{\mathrm{act}}
=
\mathcal{L}_{\mathrm{expert}}
\big(
\pi_{\mathrm{expert}}(z_v \oplus z_t),
a
\big),
\end{equation}
where $\mathcal{L}_{\mathrm{expert}}$ denotes the expert action loss. In summary, the overall objective is:
\begin{equation}
\mathcal{L}
=
\lambda_1 \mathcal{L}_{\text{soft}}
+
\lambda_2 \mathcal{L}_{\text{con}}
+
\lambda_3 \mathcal{L}_{\text{act}},
\end{equation}
where $\lambda_1$, $\lambda_2$, and $\lambda_3$ are adjustable weighting parameters.

\begin{figure*}[!t]
    \centering
    \includegraphics[width=1.0\linewidth]{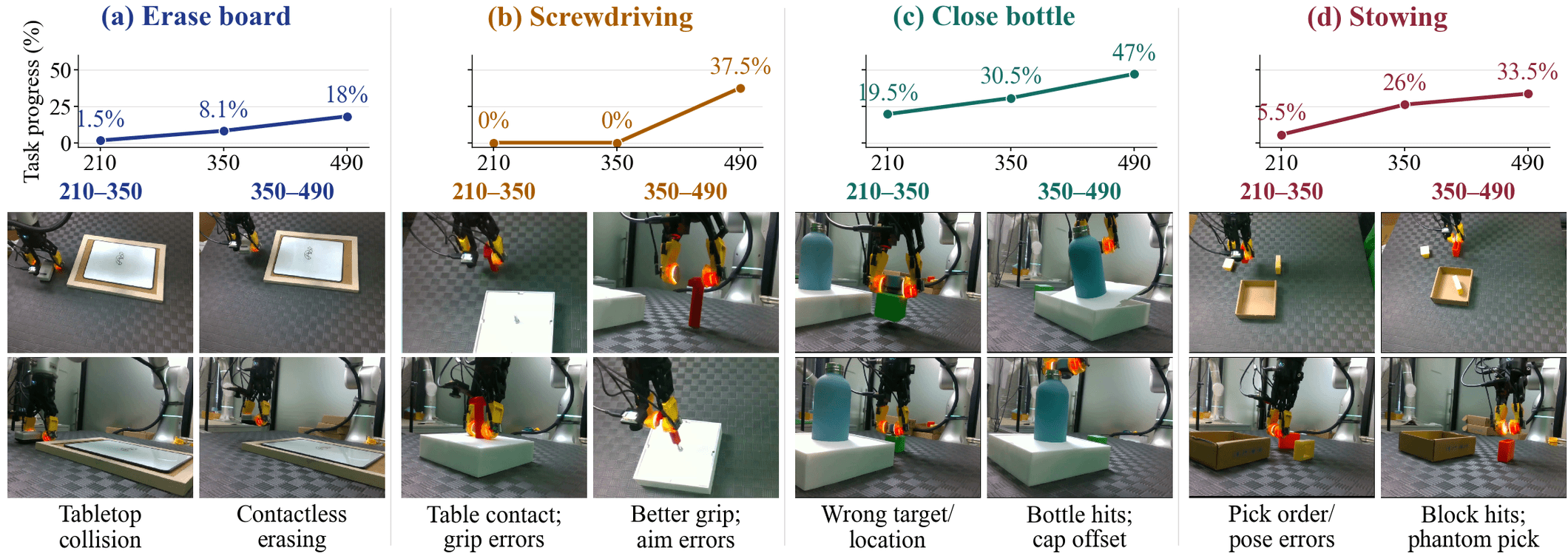}
    \caption{Post-training scaling laws across four real-world tasks. The bottom section indicates the inflection points as more data are added.}
    \label{fig:single_task_scaling_law}
\end{figure*}

\section{Experiment}

\subsection{Experiment Setup}

\paragraph{Tasks and Data Collection.} We evaluate data collection efficiency on six tasks: clamp seal, towel hanging, pick bread, erase board, pick block, and insert flower, collecting approximately 200 demonstrations per task. We then study data scaling and force–tactile learning on four more complex, contact-rich tasks: erase board, screwdriving, close bottle, and stowing. For these experiments, we collect 490 demonstrations per task across tactile fidelity levels of 100\%, 75\%, and 50\%.

\paragraph{Evaluation Methods.} We first evaluated data collection efficiency across six tasks. We then used 210, 350, and 490 demonstrations to assess post-training scaling laws. Finally, we validated ForceVT and the baselines using all demonstrations. We included the following baselines: a) VisTacLinear: this method fuses vision and tactile features using simple linear fusion. b) TacFiLM~\cite{morissette2026tactile}: this method integrates tactile tokens into the vision encoder via FiLM~\cite{perez2018film}. c) TactileConcat: as described in TacFiLM, we implemented the tactile concatenation strategy used in prior works~\cite{bi2026vla, hao2025tla, huang2025tactile, li2025adaptive, yu2024octopi, zhang2025vtla}, using T3~\cite{zhao2024transferable} as the tactile encoder.

\paragraph{Training and Inference.} Our experiments used $\pi_{0.5}$~\cite{intelligence2025pi_}, modified to learn relative poses as in UMI~\cite{chi2024universal}. We trained all models for 50,000 steps using wrist-camera images alone or paired with tactile and gripper-force signals, depending on the experimental setting. All models were evaluated on a Flexiv Rizon-4 robot arm with a Flexiv GN-01 gripper. For each model, we conducted 20 trials at five random object positions, with four trials per position, and reported the mean \textbf{task progress} following $\pi_{0.5}$~\cite{intelligence2025pi_}.
To evaluate tactile robustness, we measure the mean performance drop relative to the reference tactile fidelity:
\begin{equation}
\Delta_{\mathrm{f}}
= \frac{1}{|\mathcal{F}_{\mathrm{deg}}|}
\sum_{f \in \mathcal{F}_{\mathrm{deg}}}
\left(P_{f_{\mathrm{ref}}} - P_f\right),
\end{equation}
where $P_f$ denotes the mean task progress at fidelity $f$,
$f_{\mathrm{ref}}$ is the reference fidelity, and
$\mathcal{F}_{\mathrm{deg}}$ contains the evaluated degraded fidelities.
We use 100\% as the reference and evaluate degradation at 75\% and 50\% fidelity.

\begin{table}[!t]
\centering\captionsetup{position=top}\caption{Data collection efficiency and policy performance on vision-only tasks.
        }
        \label{tab:data_collect_efficiency} 
        \footnotesize 
        \setlength{\tabcolsep}{2pt}
        \resizebox{\linewidth}{!}{%
        \begin{tabular}{c|ccc}
        \toprule
        Task & \makecell{Number of\\demos} & \makecell{Data Collection\\Duration (min)} & \makecell{Task\\progress} \\
        \midrule
        Clamp Seal    & 229 & 28.86 & \textbf{60.0\%} \\
        Towel Hanging & 206 & 27.60 & \textbf{47.5\%}  \\
        Pick Bread    & 199 & 17.16 & \textbf{64.5\%}  \\
        Erase Board   & 215 & 26.16 & \textbf{26.0\%}  \\
        Pick Block    & 214 & 14.41 & \textbf{85.0\%}  \\
        Insert Flower & 212 & 24.72 & \textbf{22.0\%}   \\
        \bottomrule
        \end{tabular}%
        }
        
\end{table}

\subsection{Experimental Results}

\begin{table*}[!t]
\centering
\begin{minipage}[t]{0.52\textwidth}
\centering
\captionsetup{position=top}
\caption{Real-world evaluation of the vision-tactile policy across contact-rich tasks.}
\label{tab:forcevt}
\resizebox{\linewidth}{!}{
\begin{tabular}{c|cccc|c}
\toprule
Methods & Erase Board & Screwdriving & Close Bottle & Stowing & Average \\
\midrule
Vision-only & 18.0\% & 38.0\% & 47.0\% & 34.0\% & 34.3\% \\
\midrule
VisTacLinear & 31.5\% & 31.0\% & 50.5\% & 45.5\% & 39.6\% \\
TacFiLM~\cite{morissette2026tactile}
& 5.5\% & 36.5\% & 39.0\% & 43.5\% & 31.1\% \\
TactileConcat~\cite{zhao2024transferable}
& 20.0\% & 9.0\% & 22.5\% & 30.0\% & 20.4\% \\
ForceVT (Ours)
& \textbf{49.0\%} & \textbf{48.5\%} & \textbf{58.0\%}
& \textbf{55.0\%} & \textbf{52.6\%} \\
\bottomrule
\end{tabular}
}
\end{minipage}%
\hfill
\begin{minipage}[t]{0.46\textwidth}
\centering
\captionsetup{position=top}
\caption{Policy robustness on bottle closing across tactile fidelities.}
\label{tab:robustness}
\resizebox{\linewidth}{!}{
\begin{tabular}{c|ccc|cc}
\toprule
Method & 100\% & 75\% & 50\%
& Avg. $\uparrow$ & $\Delta_{\mathrm{f}}\!\downarrow$ (pp) \\
\midrule
VisTacLinear
& 50.5\% & 30.5\% & 26.5\% & 35.8\% & 22.0 \\
TacFiLM~\cite{morissette2026tactile}
& 39.0\% & 18.5\% & 14.5\% & 24.0\% & 22.5 \\
TactileConcat~\cite{zhao2024transferable}
& 22.5\% & 15.5\% & 18.0\% & 18.7\% & 5.8 \\
ForceVT (Ours)
& \textbf{58.0\%} & \textbf{57.0\%} & \textbf{53.0\%}
& \textbf{56.0\%} & \textbf{3.0} \\
\bottomrule
\end{tabular}
}
\end{minipage}
\end{table*}

\paragraph{Data Collection Efficiency} 
Figure~\ref{fig:single_generation} and Table~\ref{tab:data_collect_efficiency} present the data collection results across six tasks. Overall, pick-and-place tasks achieve over 60\% task progress with less than 20 minutes of demonstration data. Specifically, pick block achieves over 80\% task progress, while pick bread achieves approximately 60\%. More complex tasks, such as towel hanging and board erasing, achieve moderate performance with approximately 30 minutes of demonstration data. However, limited perception leads to failures such as tabletop collisions and contactless phantom erasing. The clamp seal task requires precise gripper-width control and achieves 60\% task progress. The insert flower task requires reasoning about the gripper pose, but incorrect poses can obstruct the view, making it challenging for a model using only wrist-camera images. In conclusion, our system also exhibits high data efficiency compared to exUMI~\cite{xu2025exumi}, which reports an 80\% success rate with 31 minutes of demonstration data.

\paragraph{Post-training Scaling Law} Figure~\ref{fig:single_task_scaling_law} shows that task performance improves as the number of demonstrations increases. The largest improvement occurs in screwdriving, where task progress rises from 0\% with 350 demonstrations to 37.5\% with 490 demonstrations. Erase board, close bottle, and stowing also improve, reaching 18.0\%, 47.0\%, and 33.5\%, respectively. Across these four complex, contact-rich tasks, the qualitative improvements suggest better \textbf{depth perception} and \textbf{observation–action alignment}. The former is reflected in fewer tabletop collisions and contactless erasing attempts in erase board, as well as more appropriate approach heights in close bottle and screwdriving. The latter is reflected in improved alignment during screwdriving and bottle closing and fewer block collisions during stowing.

\begin{figure*}
    \centering
    \includegraphics[width=1.0\linewidth]{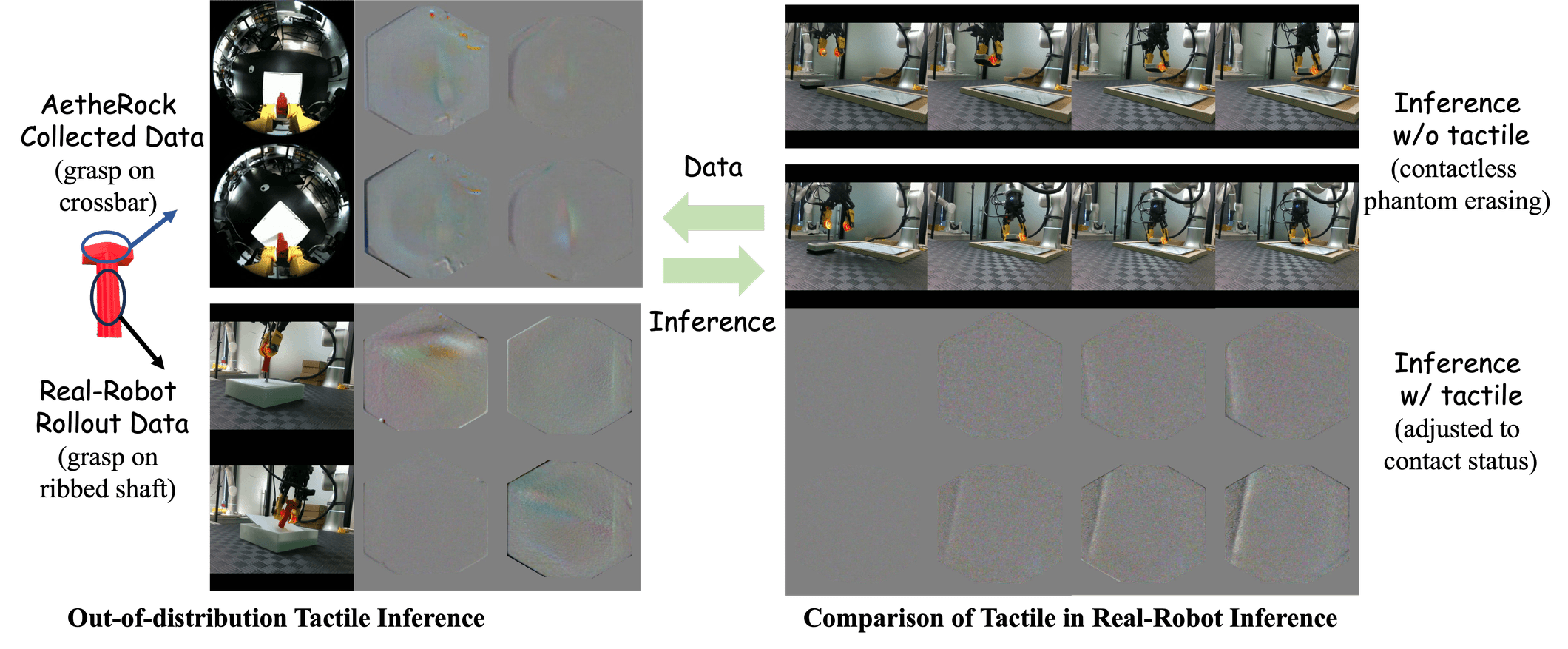}
    \caption{Visualization of tactile challenges in training and inference. The left shows grasp pose out-of-distribution during training. The right shows inference with or without tactile sensing.}
    \label{fig:tactile_result}
\end{figure*}

\paragraph{Vision-Tactile Learning}
Tables~\ref{tab:forcevt} and~\ref{tab:robustness} report vision-tactile policy performance and robustness, respectively. Overall, tactile sensing improves performance in both VisTacLinear and ForceVT. Rollout analysis reveals two key observations:
\textbf{(1) Tactile sensing indicates contact status.} As shown in the rollout examples in Figure~\ref{fig:tactile_result}, policies with tactile sensing tend to establish physical contact, whereas vision-only policies rely on visual similarity and may perform contactless phantom actions.
\textbf{(2) Tactile signals can shift out of distribution during inference.} The data examples in Figure~\ref{fig:tactile_result} show different tactile signals during training and inference in screwdriving. This discrepancy arises from inconsistent grasp positions: the gripper grasps a crossbar during training but a ribbed shaft during inference.
While TacFiLM and TactileConcat generally perform worse than simple linear fusion, suggesting that earlier vision-tactile fusion is less effective than later latent fusion in our setting. Their weaker performance may also reflect limited generalization across tactile sensors, which may contribute the absence of our sensor from T3's training data.

\textbf{Robustness to tactile degradation.} ForceVT maintains higher performance as tactile quality degrades. As shown in Table~\ref{tab:robustness}, ForceVT achieves a mean task progress of 56.0\% and an average degradation $\Delta_f$ of 3.0 in bottle closing, compared with 35.8\% and 22.0 for VisTacLinear, respectively. These results indicate that force-guided vision-tactile learning better preserves policy performance under the evaluated tactile degradation conditions.

\subsection{Ablation Study}

\begin{table}
    \centering
    \small
    \captionsetup{position=top} \caption{Ablation study of different training strategies. Our method includes a simple visuo-tactile fusion module, Tactile Random Masking, Force-Vision-Tactile Soft Alignment, and a Cross-Modality Alignment core module. We iteratively ablate each module to evaluate its importance. Parentheses indicate gains over the preceding row in percentage points.} 
    \label{tab:abstudy} \begin{tabularx}{\linewidth}{Xr} 
        \toprule
        \textbf{Method} & \makecell{\textbf{Average Task}\\\textbf{Progress (\%)}} \\
        \midrule
        Vision-only                &  47.0\% \\ 
        + ForceVision-Tactile Softalign  & 52.0\% (+5.0) \\
        + Cross-Modality Alignment  & 56.0\% (+4.0) \\
        + Tactile Random Masking     & 58.0\% (+2.0) \\
        \bottomrule
    \end{tabularx}
    
\end{table}

We ablate the core components of ForceVT: simple vision-tactile linear fusion, tactile random masking (following DINOv3~\cite{simeoni2025dinov3}), soft force-vision-tactile alignment, and cross-modality alignment. As summarized in Table~\ref{tab:abstudy}, soft force-vision-tactile alignment improves task progress by approximately 5\% over the vision-only baseline. Cross-modality alignment provides a further gain of 4\%, while tactile random masking adds approximately 2\%. These incremental gains suggest that the components provide complementary benefits to policy performance.

\section{Limitations}
A primary limitation of this work is an inconsistency in tactile contact signals between training and inference. AetheRock enables flexible manipulation and data collection, but real-robot inference, with its limited perception and execution, produces edge contacts and pad contacts. Developing algorithms and involving inference feedback in data collection to solve this generalization issue will be a focus of future work. Additionally, tactile sensing provides contact texture signals and also includes shear force perception. Assembling a robot system with a low-cost 6-axis F/T sensor, Coin-FT~\cite{choi2025coinft}, can better align with tactile signals. Finally, tactile sensors show improvement in our work with contact status guidance signals, but this is weak due to our current study concentrating on post-training. Studying pre-training with a large amount of data will be future work.

\section{Conclusion}
\label{sec:conclusion}

In this paper, we focus on hardware-algorithm co-design to propose a robotic teaching system. First, we present AetheRock, an arm-worn device for collecting gripper force, vision, and tactile data. Next, we propose ForceVT to learn robust vision-tactile representations through force guidance. We validate the data collection efficiency of AetheRock, the post-training scaling laws, and the vision-tactile learning algorithm on real-world robotic tasks. In summary, AetheRock collects robust multi-sensor data, and our algorithm improves visuo-tactile robustness.

\bibliographystyle{IEEEtran}
\bibliography{references}

\end{document}